\documentclass{article}

\usepackage[preprint]{neurips_2023}

\usepackage[utf8]{inputenc} 
\usepackage[T1]{fontenc}    
\usepackage{hyperref}       
\usepackage{url}            
\usepackage{booktabs}       
\usepackage{amsfonts}       
\usepackage{nicefrac}       
\usepackage{microtype}      
\usepackage{xcolor}         
\usepackage{graphicx}
\usepackage{amsmath}
\usepackage{color, colortbl}
\usepackage{natbib}
\definecolor{Gray}{gray}{0.9}

\newcommand{\term}[1]{\textbf{\textit{#1}}}

\title{Offline Reinforcement Learning for Optimizing Production Bidding Policies}

\author{%
  Dmytro Korenkevych \\
  \texttt{dkorenkevych@meta.com} \\
  \And
  Frank Cheng \\
  \texttt{frankcheng@meta.com} \\
   \And
  Artsiom Balakir \\
  \texttt{artsiom@meta.com} \\
   \And
  Alex Nikulkov \\
  \texttt{alexnik@meta.com} \\
   \And
  Lingnan Gao \\
  \texttt{lgao@meta.com} \\
   \And
  Zhihao Cen \\
  \texttt{zcen@meta.com} \\
   \And
  Zuobing Xu \\
  \texttt{zuobingxu@meta.com} \\
   \And
  Zheqing Zhu \\
  \texttt{billzhu@meta.com} \\
  \\
  AI at Meta \\
}

\begin{document}

\maketitle




\begin{abstract}
The online advertising market, with its thousands of auctions run per second, presents a daunting challenge for advertisers who wish to optimize their spend under a budget constraint. Thus, advertising platforms typically provide automated agents to their customers, which act on their behalf to bid for impression opportunities in real time at scale. Because these proxy agents are owned by the platform but use advertiser funds to operate, there is a strong practical need to balance reliability and explainability of the agent with optimizing power. We propose a generalizable approach to optimizing bidding policies in production environments by learning from real data using offline reinforcement learning. This approach can be used to optimize any differentiable base policy (practically, a heuristic policy based on principles which the advertiser can easily understand), and only requires data generated by the base policy itself. We use a hybrid agent architecture that combines arbitrary base policies with deep neural networks, where only the optimized base policy parameters are eventually deployed, and the neural network part is discarded after training. We demonstrate that such an architecture achieves statistically significant performance gains in both simulated and at-scale production bidding environments. Our approach does not incur additional infrastructure, safety, or explainability costs, as it directly optimizes parameters of existing production routines without replacing them with black box-style models like neural networks.

\end{abstract}

\section{Introduction} \label{sec:intro}

An important problem for advertisers participating in online ads markets is to maximize their total received value under a budget constraint \cite{perlich2012bid}. 
There are many flavors of these markets, but the most common is some kind of auction. For each available impression, these receive the bids of advertisers (together with features of the ad and ad slot) and output an allocation, specifying which advertiser wins the impression, along with a price vector, specifying how much each advertiser participant must pay. 
For example, in the widely adopted second-price auction, the price is the second-highest bid in the auction \cite{edelman2007internet, krishna2009auction}.
The sum of prices paid by an individual advertiser during a given period cannot exceed the advertiser's budget for that period; this is the advertiser's budget constraint.

Modern ad markets operate at a scale of thousands of ads auctions per second. It is impractical for advertisers to carefully evaluate all of these opportunities on an individual level using their own computational and data resources \cite{perlich2012bid, mou2022sustainable}. To alleviate this issue, many auto-bidding systems have been introduced (chief among them from platforms themselves), where the optimal bid values are evaluated by proxy agents on behalf of the advertisers \cite{hosanagar2008optimal, zhang2014optimal, cai2017real, wu2018budget, zhao2018deep, feng2018learning, mou2022sustainable, badanidiyuru2022incrementality, zhou2022multi,yuan2013real, wang2015real, wu2018budget, he2021unified}. Even for auto-bidding agents, computing the optimal bid values exactly under budget constraints in real-time is infeasible due to varied market competition and auction volumes  \cite{zhang2015statistical, cai2017real}. Hence, heuristic bidding policies are usually employed. These policies come in two flavors: handcrafted heuristic adaptive strategies, such as feedback-based control, whose policies can be easily bounded and articulated \cite{bennett1993development, perlich2012bid, zhang2014optimal, wu2018budget, yang2019bid, zhang2016feedback}, and black box strategies which are tuned by learning from data \cite{zhang2014optimal, cai2017real, wu2018budget, mou2022sustainable, zhou2022multi, he2021unified}.

Studies have shown that the second approach can significantly improve on the first \cite{zhao2018deep, cai2017real, feng2018learning, mou2022sustainable, he2021unified}, yet it has not been widely employed in consumer technology companies because advertisers strongly prefer using agents that are safe and explainable due to the enormous financial impact of making an error. This preference is shared by platforms, as they share the significant financial consequences of mistakes and must also work to fix mistakes efficiently. 
 
In this paper, we build agents bridging these gaps using \term{offline reinforcement learning}. 
Our approach is modular with respect to a \term{base policy}, which is an explainable parameterized heuristic adaptive strategy that enjoys wide acceptance by advertisers. A simple PID controller \cite{bennett1993development} is one example of such a base policy. These base policies are highly restricted in both their action space as well as their observation space. Our output is an optimized policy that shares the parameter space of the base policy. 
Empirically, we show that our approach improves on a highly tuned base policy in a real production system.
This allows us to adhere to advertisers' and platforms' transparency requirements while delivering more total value to advertisers.
Under a reinforcement learning framework, a bidding problem can be formulated as an episodic Markov Decision Process (MDP), where each ad campaign represents one episode, and the whole market (users and competing ads) comprise the environment. The bidding policy makes decisions sequentially based on the campaign state, and these decisions affect future states and long-term outcomes. Given this MDP model, the RL framework allows for maximization of long-term outcomes through learning optimized bidding policies from real data. Conventionally, modern RL algorithms are applied to train parameters of powerful function approximators, namely deep neural networks \cite{andrychowicz2020learning, haarnoja2018soft, kalashnikov2018scalable, levine2016end, mahmood2018benchmarking, mnih2016asynchronous, korenkevych2019autoregressive}. In this work, we follow a hybrid approach where we combine an arbitrarily parameterized base bidding policy, whose output is differentiable with respect to its parameters, with deep neural networks to optimize the policy parameters. The optimized bidding policy can be deployed in production without incurring additional infrastructure or safety costs. Namely, we propose to use an actor-critic agent architecture \cite{konda1999actor}, where the actor follows a Gaussian stochastic policy with the mean parameterized with the production bidding policy and the variance parameterized with a deep neural network. The critic model is parameterized with a deep neural network as well. The bidding policy parameters are initialized at the default production values. We train the agent completely offline by applying an offline RL learning algorithm, Conservative Q-Learning (CQL) \cite{kumar2020conservative}. The training data is collected in production by a default production policy with a small amount of exploration noise. Therefore, our agent's policy is initialized as the behavior policy at the beginning of training, which helps to mitigate some of the challenges of offline RL, such as actions' distributional shift between the behavior policy and the learned policy \cite{kumar2019stabilizing, levine2020offline, kakade2002approximately, fujimoto2019off, eysenbach2023connection}.

To demonstrate the merit of the described approach empirically, we apply it to the optimization of the parameters of a real-time bidding policy. We first validate our approach in a high-fidelity bidding simulator and show that it can consistently improve upon the baseline policy and produce statistically significant performance gains. Most importantly, we apply the proposed approach to a real production system and optimize production bidding policy parameters using real logged data. In a series of large-scale A/B tests, we report statistically significant gains of an optimized model compared to the original baseline. 

These results confirm that the described approach of using offline RL with a hybrid agent architecture can be successfully applied to optimizing varied control policies in real production systems. We note that this approach is general with respect to the base policy, and although we empirically validate it in the auto-bidding setting, it can be applied to any control policy as long as the appropriate training data can be collected. 

\section{Background and Related Work}
\subsection{Reinforcement Learning}
Reinforcement learning (RL) framework models an agent interacting with an environment at discrete time steps \cite{sutton1998introduction}. At each step $t$ the environment presents the agent with the environment state $S_t \in \mathcal{S}$ and a scalar reward signal $R_{t} \in R$. In response, the agent chooses an action $A_t \in \mathcal{A}$ according to a policy defined by a probability distribution function $\pi(a|s) = P\left\{ A_t=a | S_t=s \right\}$. At the subsequent time step $t + 1$ partially due to the agent's action, the environment transitions to a new state $S_{t+1}$ and produces a new reward value $R_{t+1}$ based on a transition probability distribution function $p(s^{\prime}, r|s, a) = \Pr\left\{ S_{t+1} = s^{\prime}, R_{t+1}=r | S_t = s, A_t = a \right\}$.
The objective of the agent is to maximize the expected return defined as the discounted sum of future rewards $G_t = \sum_{k=t}^\infty \gamma^{k-t} R_{k+1}$, where $\gamma\in[0,1]$ is a discount factor.
In practice, the agent usually does not have access to the environment state $s_t$ but observes it partially through an observation vector $O_t$.

Given a policy $\pi$, a value function of the policy is defined as: 
$$V_{\pi}(s) = \mathbb{E_\pi} [G_t | S_t = s].$$
Similarly, a state-action value function, also called $Q$-function, is defined as:
$$Q_\pi(s, a) = \mathbb{E_\pi} [G_t | S_t = s, A_t = a].$$

\subsection{Related Work}

Our work is related to literature that models the bidding problem as an MDP and solves it using learning approaches \cite{weed2016online, feng2018learning, balseiro2019contextual, amin2012budget, cai2017real, wu2018budget, mou2022sustainable, wang2023adversarial}. Many of these approaches, however, formulated the problem as contextual bandits \cite{weed2016online, feng2018learning, balseiro2019contextual}, which does not account for intra-episode dynamics. Other proposed solutions are not applicable at large production scales \cite{amin2012budget} or require replacing the production bidding policy with a particular form of the policy function, usually a deep neural network, which incurs additional infrastructure, support, and safety costs \cite{amin2012budget, cai2017real, wu2018budget, mou2022sustainable, wang2023adversarial, zhou2022multi}. 
In \cite{cai2017real} the authors propose applying a dynamic programming approach to derive an optimal policy for an individual auction. The policy computation requires access to the predicted click-through rate (pCTR) for each individual auction. While the approach is suitable for small-scale problems, it is not applicable to real production scenarios.
In \cite{wu2018budget} the authors applied the DQN algorithm to collect data and train an approximate Q-function parameterized by a deep neural network. The bidding policy in this case is an epsilon-greedy policy derived from the Q-function, which is a black box to the advertiser. This would incur additional costs and risks to implement in the production environment. 

In \cite{mou2022sustainable} the authors proposed applying online reinforcement learning to learn the bidding policy directly in a production environment. To mitigate risks during data collection, the authors designed a safe exploration policy, which utilized both a learned policy and a safe prior production base policy. Under this online framework, the authors demonstrated iterative improvement of the learned policy performance after each model update. While this is an exciting result that shows interesting prospects of employing online RL agents to control ad campaign budgets, implementing such an iterative online learning framework still remains a challenge in most production environments. In particular, the learned policy is based on deep neural networks and therefore requires the deployment of a neural network in production for data collection. Additionally, computing the policy output at each time step in their approach requires multiple samples from a uniform distribution to be passed through a Q-function neural network, which incurs significant computational costs at a large scale. In contrast, our approach requires a single pass of a base production policy that, by design, is already supported and does not incur additional costs.  

Another relevant line of work applied RL to tuning simple parameterized policies, such as PID controllers \cite{dogru2022reinforcement, shipman2021learning, carlucho2017incremental, sun2021design, sedighizadeh2008adaptive}. In \cite{dogru2022reinforcement} the authors framed the problem as a contextual bandit problem, where the actions represent PID controller parameters, and an entire episode outcome is treated as a single reward value. In the context of bidding where the episodes may contain thousands of time steps, this approach would be sample inefficient and impacted by a high variance in returns, as an entire episode provides a single data point in this case. Similarly, in \cite{shipman2021learning} an RL agent manipulated the parameters of a PID controller with the agent's actions mapped to the controller parameters values. In \cite{carlucho2017incremental} the authors applied tabular $Q$-learning to output the PI controller parameters to control a mobile robot. In \cite{sun2021design, sedighizadeh2008adaptive} the authors applied online RL algorithms to train agents that output PID controller parameters in real time. In real production systems, changing the base policy parameters dynamically at every time step may require additional infrastructure investment on one hand, and may increase safety risks and policy explainability costs on the other. In contrast, we learn from individual time steps in each episode while at the same time we treat base policy parameters as learnable parameters that do not change once the model is deployed.


\section{Methods}

\subsection{Problem Formulation}
\label{sec:problem_formulation}
In this section, we formulate ad campaign budget control as a sequential decision-making problem in the context of reinforcement learning. The campaign budget control agent is given two constraints: duration of budget availability, and the total amount of budget the agent is not allowed to spend over for the duration. At every time step, the agent bids a portion of its budget to participate in a set of available auctions. In our system, the auto-bidding policy output gets scaled by the expected conversion rate (eCVR) before participating in an auction. The same policy output value is used in all auctions the agent participates in within a single time step, scaled by the corresponding eCVR in each auction. In a real system, a campaign may participate in hundreds or even thousands of auctions in a single time step. In our Markov decision process (MDP) model, we treat the bidding policy output as the agent's action, and the scaling by eCVR as a part of the environment mechanics, which is inaccessible to the agent. 
The agent's budget is then adjusted accordingly based on the agent's action and a set of auctions the agent has won. 
If the agent wins an auction, the agent receives the auction target, such as an impression. 
The goal of the agent is to maximize the cumulative value of all auction targets it wins throughout the duration of budget availability without overspending the budget. 

To formally model this process as an MDP, we define interactions between the agent and the auction environment in terms of a quintuple $\mathcal{E} = (\mathcal{S}, \mathcal{A}, \mathcal{R}, T, \rho)$: 
\begin{enumerate}
\item State space $\mathcal{S}$: at each time step $t$ the state element $S_t \in \mathcal{S}$ describes the full Markovian state of the environment, which includes the state of the campaign controlled by the agent, the set of auctions the agent is eligible to participate in at the given time step, and the states of all competing ads campaigns in those auctions. In practice, the agent doesn't have access to the majority of this information and instead operates based on a limited observation vector $O_t \in O$ that mainly contains the information about the campaign controlled by the agent. 
\item Action space $\mathcal{A}$: The agent executes actions that correspond to auction bids $A_t \in \mathbb{R}$. The same value $A_t$ (with subsequent scaling by eCVR) is used in all auctions the campaign participates in at time step $t$.  
\item Reward space $\mathcal{R}$: We define the reward $R_t \in \mathcal{R}$ to be equal to the aggregated value of outcomes of all auctions the agent's campaign participated in at the time step $t$: $R_t = \sum\limits_{i=1}^{N_t} \delta^i_t W^i_t$, where $N_t$ is the number of auctions the agent participates in at the step $t$, $W^i_t \in \{0, 1\}$ is the outcome of the $i$-th auction, i.e. win or loss, $\delta^i_t$ is the value of winning the $i$-th auction, e.g. the value of a conversion. 
\item Auction campaign horizon $T$: given that auction campaign 
budgets usually have an expiration date, we set an auction campaign horizon $T$ for the auction environment. We consider each individual ad campaign as one realization, or one episode, of the above MDP.
\item State transition probability $\rho$: given the state and action pair $(S_t, A_t)$, the environment transitions to the next state $S_{t+1}$ according to the conditional probability distribution $\rho(\cdot \mid S_t, A_t)$. While the exact form of $\rho$ is complex, essentially it models auction mechanisms and bidding policies of all competing ads campaigns at each time step.  
\end{enumerate}

\subsection{Offline reinforcement learning for tuning production bidding policies.}

In this work, we view production bidding heuristics as trainable RL policies. Let us consider a production bidding function $F_w: S \subset R^d \rightarrow A \subset R^1$ parameterized by a vector $w$, which maps certain ad campaign state representations $s \in S$ to scalar auction bids. In the proposed view, $s$ is an environment state representation, $a = F_w(s)$ is an action, and the vector $w$ represents learnable policy parameters. If the function $F_w$ is differentiable with respect to its parameters $w$, we can use reinforcement learning and gradient descent methods to optimize its parameters from real data.

As an example, consider a simple proportional and integral (PI) controller \cite{bennett1993development}, defined as 
$$a_t = a_{t - 1} + K_p e_t + K_i \sum_{\tau = 0}^te_\tau,$$
where $a_t$ is the controller output at time $t$, $e_t$ is some error measure at time $t$ that we aim to minimize (for example, a discrepancy between the fraction of campaign budget spent and the estimated consumed fraction of total impression opportunities), $K_p$ and $K_i$ are the proportional and integral coefficients respectively. In this modeling approach, the vector $$s_t = [a_{t-1}, e_t, \sum_{\tau = 0}^te_\tau]$$ is a state representation vector at time $t$, the vector $$w = [K_p, K_i]$$ is a vector of learnable parameters, and the function $$F_w(s_t) = a_{t - 1} + K_p e_t + K_i \sum_{\tau = 0}^te_\tau$$ is a deterministic policy. 
Please note that the above PI controller is merely an example of a base policy and it does not represent the policy that runs in our production system and that we optimized in this work. 

It is unsafe to deploy an under-trained agent in a production environment, even on a small fraction of traffic. At certain production scales, even a slightly sub-optimal policy can incur the company millions of dollars in losses within hours, in addition to customers' dissatisfaction with the campaigns' performance. For this reason, we opted to train the agent offline following an offline RL approach. Offline RL attempts to learn a high-performing policy from the data collected by a separate data generation process (also called behavior policy). In offline RL setting the agent does not interact with the actual environment until fully trained, which is an important enabling factor in performance-critical production environments.

\begin{figure}[tb]
  \centering
  \includegraphics[width=1.0\linewidth]{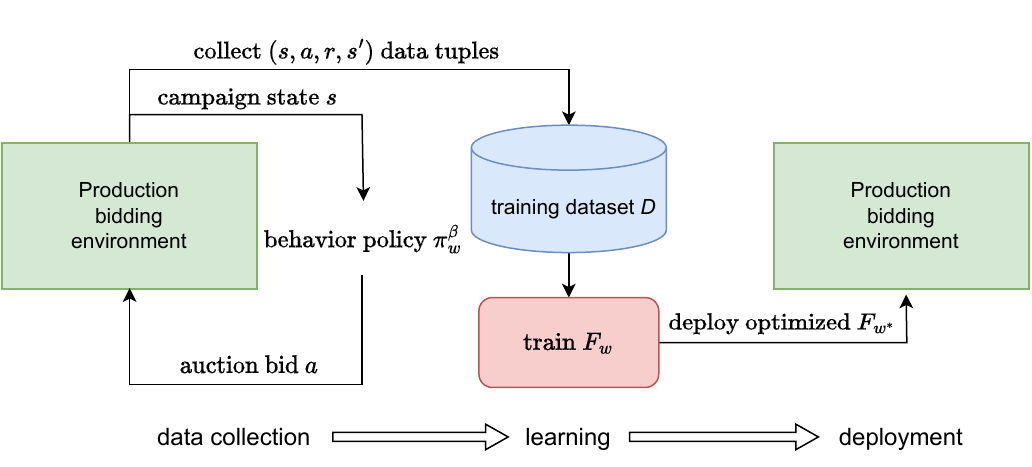}
  \caption{The flowchart of the proposed approach. Here $F_w$ denotes the base production bidding policy, parameterized by the vector $w$; $\pi^\beta_w$ denotes the base policy with default parameters and added exploration noise, defined in equation (\ref{eq:exp_noise}); $F_{w^*}$ denotes the base policy after training with optimized parameters vector $w^*$.}
  \label{fig:flowchart}
\end{figure}

Given a production base policy $F_w$, our approach consists of the following steps (Figure \ref{fig:flowchart}). First, we defined a \term{behavior policy} $\pi^\beta_w$ as the base policy with default production parameter values and a small scale Gaussian exploration noise (see Section \ref{sec:behavior_policy} for details). 
Next, we deployed the behavior policy  $\pi^\beta_w$ to the production environment and collected a training dataset $D$. We designed an agent (Section \ref{sec:agent_arch}) that combines the base policy $F_w$ and deep neural networks and applied a state-of-the-art offline reinforcement learning algorithm to optimize the agent's policy on the dataset $D$. Finally, the base policy with optimized parameters $F_{w^*}$ is deployed to production for A/B testing. The next sections describe each of the components of our approach in detail.

\subsection{Agent Architecture}
\label{sec:agent_arch}

\begin{figure}[tb]
\centering
\includegraphics[width=1.0\linewidth]{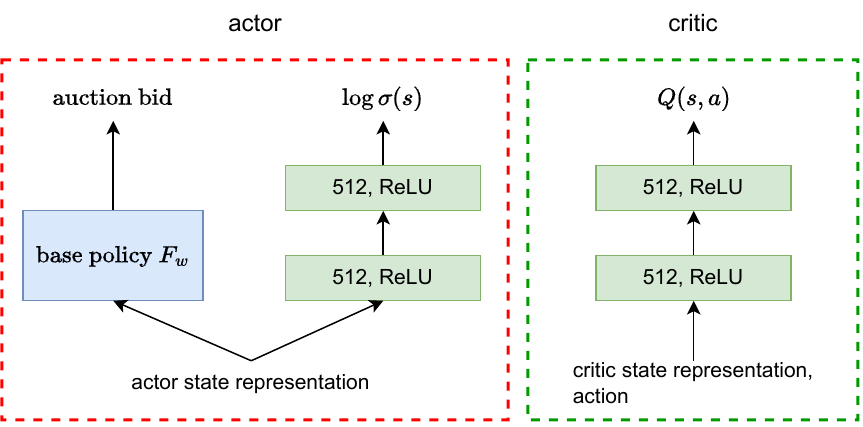}
\caption{The proposed actor-critic agent architecture. The actor consists of two components: a deterministic production base policy that parameterizes the agent's policy mean, and an MLP neural network that parameterizes the policy variance. The critic is parameterized by a separate MLP neural network.}
\label{fig:agent_arch}
\end{figure}

In order to adapt a more conventional RL setting and to be able to use state-of-the-art reinforcement learning algorithms, we designed our agent based on an actor-critic architecture (see Figure \ref{fig:agent_arch}). The actor model represents a stochastic bidding policy and defines an action distribution $\pi(\cdot | S_t)$ conditioned on the state representation vector $S_t$. The critic model approximates the value function $Q(S_t, A_t)$ conditioned on a state-action pair $(S_t, A_t)$. 
To convert a deterministic base policy $F_w$ to a stochastic policy, we defined the agent's policy as a Gaussian distribution with a mean parameterized by $F_w$, and a variance parameterized by a neural network:
\begin{equation}
\pi_{w, \phi}(\cdot | S_t) = \mathbb{N}(F_w(S_t), \sigma^2_\phi(S_t)),
\end{equation}
where $w$ is a vector of parameters of the base policy, and $\phi$ is a vector of weights of the neural network representing the variance. 

Once the agent is trained, only the base policy with optimized parameters $F_{w^*}$ is deployed to production (Figure \ref{fig:flowchart}). 
This design allows to avoid additional infrastructure and implementation costs and mitigates the safety risks of deploying a black box type model such as a neural network, to a production system. In this design, the actor is restricted to using the state representation features that are accepted by the base policy $F_w$. The critic, however, is not restricted by production constraints, since it is parameterized by a neural network and is not deployed to production. Consequently, the critic can use a larger set of input features compared to the actor, and we have expanded the state representation features set for the critic to increase the accuracy of critic predictions. A more accurate critic model results in more accurate gradient estimates with respect to the policy parameters. Therefore, we used different state representations for the actor and the critic models. In practice, we used two-layer MLP neural networks with 512 neurons in each hidden layer and ReLU non-linearities to represent both, the policy variance $\sigma^2_\phi$ and the critic models (Figure \ref{fig:agent_arch}). 

\subsection{Behavior Policy}
\label{sec:behavior_policy}
Reinforcement learning requires exploration in order to be able to discover new policies. In order to facilitate such exploration, we added a truncated multiplicative Gaussian noise to the deterministic base policy $F_w(S_t)$. Our behavior policy $\pi^\beta_w$ is therefore defined according to:
\begin{align} \label{eq:exp_noise}
\pi^\beta_w(S_t) &= F_w(S_t) \cdot (1 + \text{clip}(\varepsilon_t, -0.5, 0.5)), \\
\varepsilon_t &\sim \mathbb{N}(0, \sigma^2_{\beta}), \notag
\end{align}
where $\sigma^2_{\beta}$ is the distribution variance. The reason for truncating the noise is to exclude extreme bid values that could have a strong negative impact on a production system.  
The multiplicative noise of the form presented above is equivalent to the (truncated) conventional additive Gaussian noise with the variance scaled by the policy mean $F_w(S_t)$:
\begin{equation} \label{eq:behavior_policy}
   \pi^\beta_w(S_t) = \mathbb{N}(F_w(S_t), F_w(S_t) \cdot \sigma^2_{\beta}) 
\end{equation}
The rationale for this parameterization was that the appropriate bid values, depending on factors like the campaign's remaining budget and remaining opportunity size, can be widely different in scale, varying by several orders of magnitude. Additive exploration noise with a fixed variance would be too disruptive in states with small bids, or too negligible in states with large bids. Scaling the variance by the value of the bid itself ensures that we explore in a reasonable region relative to the bid value.

\subsection{Learning Algorithm}
\label{sec:learning_algorithm}
In this work, we used a modified version of a Conservative Q-Learning (CQL) algorithm \cite{kumar2020conservative}, which is currently one of the state-of-the-art offline RL algorithms. The general idea of the algorithm is to penalize the Q-function values on out-of-distribution (OOD) actions to negate the over-estimation bias, therefore removing the incentive for the policy to learn to favor such actions. For continuous control domains, such as in bidding, the CQL algorithm is based on a Soft Actor-Critic (SAC) algorithm, which is one of the state-of-the-art in online RL \cite{haarnoja2018soft}. Below we provide brief introductions to both of these algorithms and describe the changes we've made to adapt CQL to the bidding setting.

\paragraph{Soft Actor-Critic (SAC)} The SAC algorithm \cite{haarnoja2018soft} is based on the idea of entropy-regularized Reinforcement Learning where the standard RL objective of maximizing cumulative reward is augmented with an entropy term:
\begin{equation}
\pi^* = \arg \max_\pi \sum_t \mathbb{E}_{(s_t, a_t) \sim \rho_\pi} \left[r(s_t, a_t) + \alpha H(\pi(\cdot | s_t))\right],
\end{equation}
where $H(\pi(\cdot | s_t))$ is an entropy of the policy $\pi$, $\alpha$ is a "temperature" parameter that controls the relative weight of the entropy term in the objective, $\rho_\pi$ is a trajectory distribution induced by the policy $\pi$. This objective incentivizes the policy to explore more broadly and is better suited for learning multi-modal distributions where multiple actions have similar values. 

SAC is based on an actor-critic architecture where the actor model learns the policy and the critic model learns to approximate the corresponding Q-function $Q_\pi(s, a)$. More specifically, the critic model is trained by minimizing the soft Bellman operator error
\begin{equation} \label{sac_critic}
\begin{split}
J_Q(\theta)  = \frac{1}{2} \mathbb{E}_{(s, a) \in D} &\bigl[ (Q_\theta (s, a)\\ 
&- (r(s, a) + \gamma E_{s' \sim p} V_\theta(s')))^2 \bigr],
\end{split}
\end{equation}
where $\theta$ are the critic parameters, typically the weights in a neural network, $\gamma$ is a discount factor, $p$ is an environment state transition probability, $V_\theta(s)$ is a state-value function, $D$ is a replay buffer. 

The actor model is trained by minimizing the KL divergence between the policy and the exponential of the Q-function, which reduces to the following objective
\begin{equation}
J_\pi(\phi) = \mathbb{E}_{s \sim D} \left[\mathbb{E}_{a \sim \pi_\phi} [\alpha \log (\pi_\phi(a | s)) - Q_\theta(s, a)] \right],
\end{equation}
where $\phi$ are the policy parameters, $D$ is a replay buffer. 
\paragraph{Conservative Q-learning (CQL)}
CQL($H$) variant \cite{kumar2020conservative} of the CQL algorithm extends the SAC loss (\ref{sac_critic}) to the offline setting by adding a penalty term to the Bellman operator error objective:
\begin{equation}
\begin{split}
\min_Q\ &{\color{red}\alpha \mathbb{E}_{s \sim D}\left[\log\sum_a \exp (Q(s, a)) - E_{a \sim \hat{\pi}_\beta (a|s)}[Q(s, a)]\right]} \\
&+ \frac{1}{2} \mathbb{E}_{(s, a) \in D} \bigl[ (Q_\theta (s, a) 
- (r(s, a) + \gamma E_{s' \sim p} V_\theta(s')))^2 \bigr],
\end{split}
\end{equation}
where $D$ is an offline dataset, $\hat{\pi}_\beta$ is a behavior policy that generated the data. The terms in red highlight the penalty terms introduced in the CQL algorithm. 

The first term in the penalty penalizes the Q-values of all actions in the action space, while the second term counters the effect of the penalty on the in-distribution actions sampled from the behavior policy. The combination of these two terms, therefore, attempts to penalize the Q-values of out-of-distribution actions. In the original algorithm, the first penalty term
\begin{equation} \label{first_pen_term}
\mathbb{E}_{s \sim D}[\log\sum_a \exp (Q(s, a))]
\end{equation}
is estimated by sampling actions from a uniform distribution over the action space, and the second term
\begin{equation} \label{second_pen_term}
\mathbb{E}_{s \sim D, a \sim \hat{\pi}_\beta (\cdot|s)}[Q(s, a)]
\end{equation}
is estimated by using actions recorded in the dataset.

To achieve stable learning performance with the CQL algorithm in the bidding environment, we added several modifications to the original practical algorithm proposed in \cite{kumar2020conservative}, which we describe below. 


\begin{figure*}[tbh]
  \centering
  \includegraphics[width=1.0\linewidth]{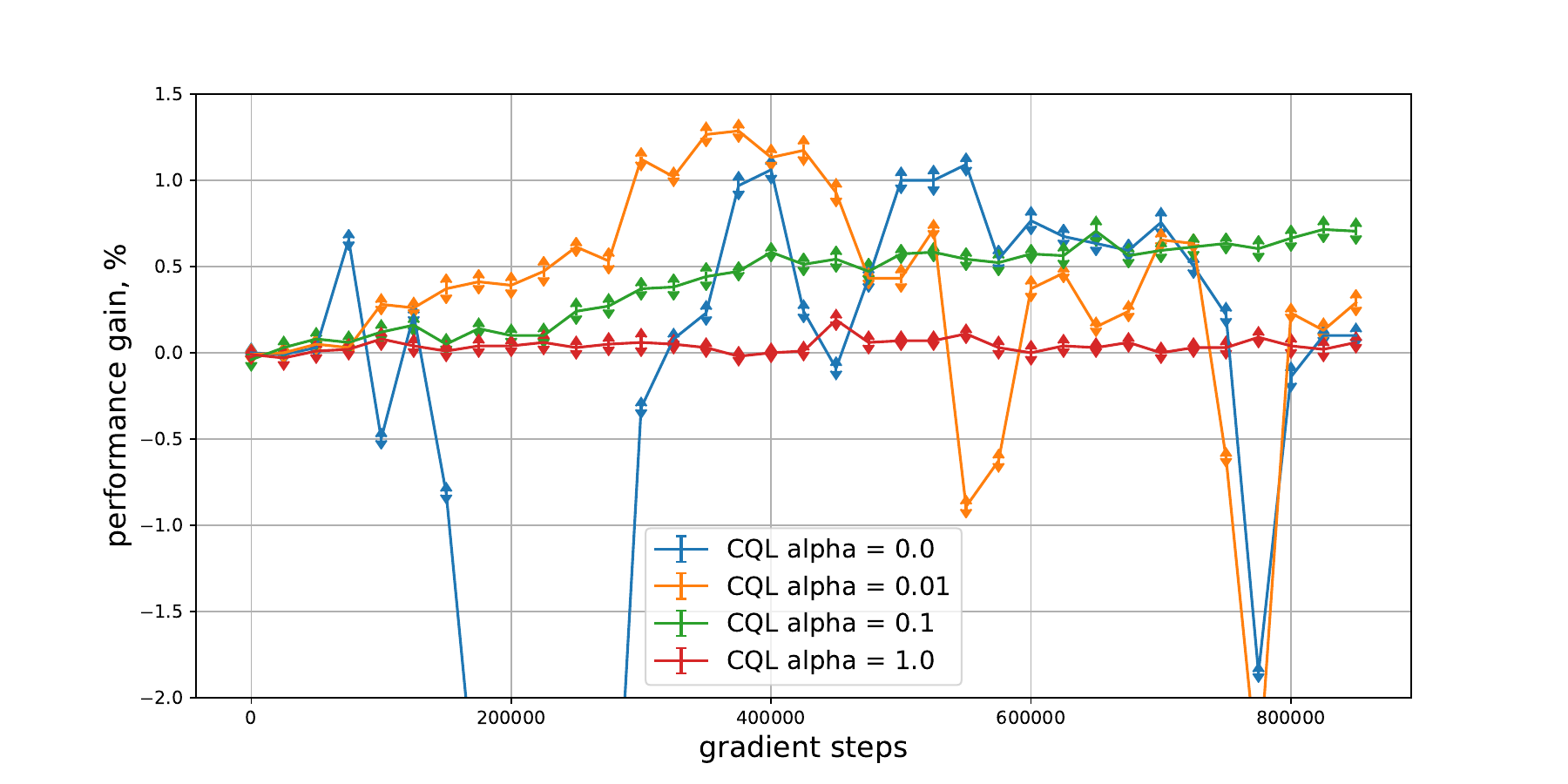}
  \caption{CQL agent learning curves in the simulator at different values of CQL penalty weight parameter. Error bars represent 95$\%$ confidence intervals computed from 1,000 evaluation episodes at each checkpoint.}
  \label{fig:learning_curves}
\end{figure*}

\begin{enumerate}

\item The first modification we made is in estimating the first penalty term (\ref{first_pen_term}). In bidding, the optimal action values in different states can differ by several orders of magnitude (since the bids are strongly influenced by campaign budgets), and in order to perform well, both, behavior and learned policy follow distributions that by design are quite narrow around these optimal actions (see e.g. (\ref{eq:exp_noise}) for details). In this setting, sampling actions uniformly across the entire action space in each state and using them to evaluate the penalty is wasteful, as the vast majority of those actions will never be taken in that state. Instead, we sample from the state-dependent uniform distribution over the interval
\begin{equation}\label{cql_sampling_interval}
I_\beta(s) = [(1 - \varepsilon) F_w(s), (1 + \varepsilon) F_w(s)],
\end{equation}
where, by design, $F_w(s)$ is the deterministic mean of the behavior policy. The parameter $\varepsilon$ was chosen to match the clipping parameter in the multiplicative exploration noise applied to the behavior policy during the data collection (\ref{eq:exp_noise}). Sampling from (\ref{cql_sampling_interval}) ensures that sampled actions stay in the relevant interval for each state. The $I_\beta(s)$ interval covers all actions produced by the behavior policy at each state, and hence we can think of it as representing an entire (state-dependent) action space. 

\item The second modification we made is in estimating the second penalty term (\ref{second_pen_term}). In our case, we have access to the behavior policy that generated the data  (i.e. the default production bidding policy with exploration noise), which often is not available in offline RL setting. Instead of using actions recorded in the data, we can leverage the access to the behavior policy and sample $K$ actions at each state $s$ directly from  
$\pi_\beta(\cdot | s)$.
Using these sampled actions to estimate the expectation in (\ref{second_pen_term}) reduces the variance of the estimate, and also reduces the risk of overfitting from re-using the actions from the dataset. 

\item We removed the entropy bonus term from the value function objective. The entropy bonus is useful in the online RL setting, as it promotes exploration. In our setting, however, the agent is trained entirely offline from the data collected by a fixed behavior policy, and maximizing policy entropy in this case is not meaningful. 

\item Since we initialize the agent to match the base policy used to collect the data $\pi^\beta_w(\cdot | s)$, we can pre-train the critic network from the collected data to have critic predictions match the actor performance at the start of the training process. Specifically, we pre-train the critic for 300,000 gradient steps before starting to train the actor. Empirically, we found that the critic loss largely converges at around that point in training. The resulting critic model matches better the true policy $Q$-function at the beginning of training and allows to compute more accurate gradients with respect to policy parameters.
 \end{enumerate}

\section{Experimental results}
To validate our approach, we performed a series of empirical experiments in both, a simulated bidding environment, and a real production system. We defined the base policy $F_w$ to be the heuristic bidding policy that is running in our production system. 
In a nutshell, it is a piece-wise polynomial function in input features with a couple dozen scalar parameters. 
We initialized our agent actor parameters to default production values. We used a behavior policy defined in (\ref{eq:behavior_policy}) to collect the training dataset for subsequent offline learning. We have experimented with various values of the variance $\sigma^2_{\beta}$ and found that the value $\sigma_\beta = 0.05$ resulted in $\sim 0.5\%$ drop in performance compared to the default production policy when deployed online. This was close to the limit at which it was still acceptable to collect the data over the course of two weeks even on a fraction of traffic. A larger variance would incur a loss in performance that wouldn't be acceptable for collecting a substantial amount of data.

\subsection{Simulation setup}

To validate the approach and perform a hyper-parameters search, we trained our bidding agent in a campaign bidding simulator. The simulator was developed internally for testing new variants of production bidding policies, and therefore it has a reasonably high fidelity with respect to the real environment. The simulator allows us to customize campaign configurations, such as budget size, campaign duration, audience size, opportunities distribution over the campaign duration, etc.

We designed a set of 100 campaign configurations with randomized opportunity distribution density functions, audience sizes, and budget sizes. The audience size determines the total number of opportunities (simulated users) available to a given campaign throughout its lifetime, and the budget determines the total amount of money a campaign can spend in one episode. Each campaign runs for one day of simulated time, and the bidding agent makes decisions every simulated minute. An episode is terminated when either the campaign spends all its budget, or it runs for a full day of simulated time. Each episode, therefore, consists of up to 1,440 time steps. At the beginning of each episode, a campaign configuration is selected from the available configurations set at random, and the environment state is initialized with the corresponding parameters. A single agent is trained on the data from all available configurations. At each time step, the campaign participates in simulated auctions where it competes over impressions against other simulated campaigns. 
We have defined the total value of simulated conversions, received at each time step, as a reward function in this simulated environment. To train the agent, we have generated an offline dataset containing $\sim 400,000$ episodes ($\sim 600$ million time steps) by running the base policy with Gaussian exploration noise in the simulator (see Section \ref{sec:behavior_policy} for the details on the behavior policy we used to collect the data). 

A major challenge in applying offline RL to production systems is that we do not have the means to evaluate the agent apart from running A/B tests in production, which are costly and time-consuming. In particular, we do not have the option to pick the best model checkpoint across the training run, since we do not know which checkpoint is the best. Consequently, we don't know for how many gradient steps we should train the model before committing it to an A/B test. Hence, the main questions we wanted to answer with our simulated experiments were:
\begin{itemize}
\item Can our approach improve upon the base policy?
\item How consistent and stable is the learning process? 
\item Can we pick reliably a model checkpoint with a positive performance gain compared to the base policy without evaluating multiple checkpoints in expensive A/B tests?
\end{itemize}
To answer these questions, we trained a set of agents with varied values of CQL penalty weight $\alpha$ in the simulator and evaluated each agent at multiple checkpoints during the training process by running 10 evaluation episodes per each of 100 campaign configurations (1000 episodes in total per checkpoint). The value of the penalty weight parameter determines how conservative the learning process is in terms of deviating from the behavior policy, and it directly affects its stability. Figure \ref{fig:learning_curves} shows the resulting learning curves. At small values of $\alpha$, the penalty term was not sufficient to mitigate the over-estimation bias on out-of-distribution actions, and it resulted in unstable learning and abrupt changes in performance. In contrast, a large value of $\alpha = 1.0$ was too restrictive and resulted in very marginal changes in the policy compared to the baseline. At the intermediate value of $\alpha=0.1$, however, the agent produced a consistent stable performance, exhibiting positive performance gains along the entire learning process. As can be seen from Figure \ref{fig:learning_curves}, after 400,000 gradient steps the performance gain of the $\alpha = 0.1$ agent consistently stays above $+0.5\%$ mark, and any model checkpoint past that point would be a suitable candidate for an A/B test. We used the value $\alpha=0.1$ in our subsequent experiments in both, simulation and a real production environment.

\begin{figure}[tbh]
  \centering
  \includegraphics[width=0.9\linewidth]{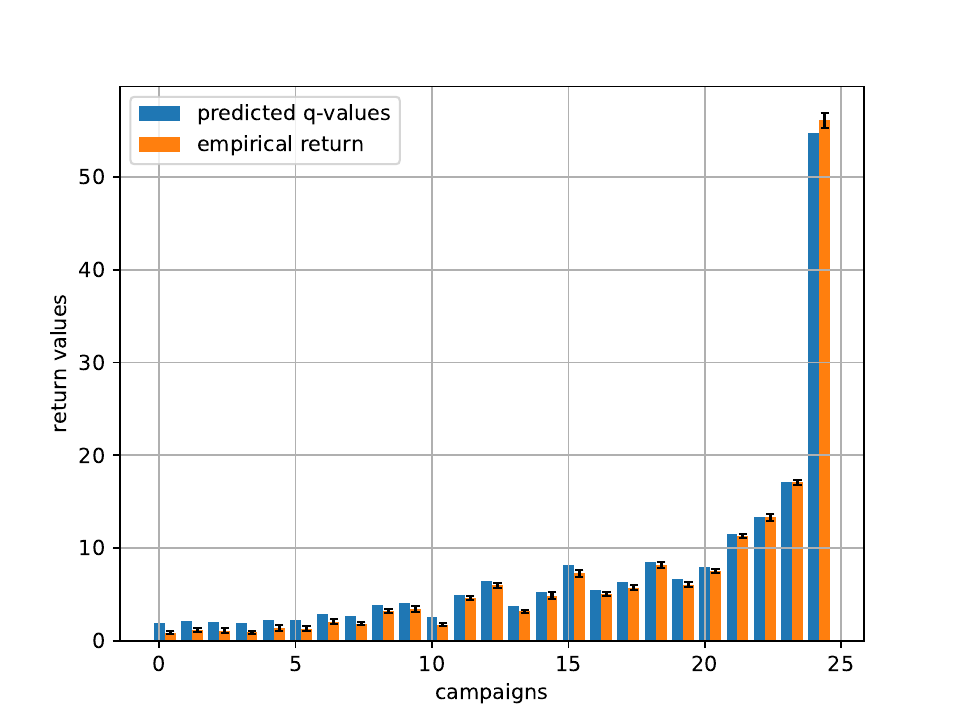}
  \caption{Learned $Q$-function predictions at the start of an episode against empirical discounted returns across 25 different campaign configurations. To estimate empirical returns, we ran the trained policy in the environment for 30 episodes for each campaign and report the mean return values. The error bars represent standard deviations. The campaigns are ordered left to right based on the campaign budget.}
  \label{fig:q_vals_campaigns}
\end{figure}
In the bidding environment, the observation data contains only highly aggregated high-level information about the environment states, such as total fraction of budget spent, total fraction of opportunities passed, average past bid value, etc. We wanted to verify that the neural network-based critic model in our hybrid agent architecture is able to predict the returns with a reasonable accuracy based on this information. For a set of 25 different campaign configurations with varied budget sizes, we generated new episodes with the trained agent and computed critic predictions at the initial state  $(s_0, a \sim \pi_{w, \phi}(\cdot, s_0))$ in each episode. We compared these predictions with the total empirical discounted returns in those episodes. As can be seen in Figure \ref{fig:q_vals_campaigns}, although the critic tends to over-estimate the returns in some cases, particularly on small budget campaigns, in general, the critic predictions are quite accurate and are able to capture the difference in performance between campaigns with different budget values. Note, that since new episodes were generated for this experiment, the data used to compute results on the figure was not part of the training dataset.

\begin{figure}[b]
  \centering
  \includegraphics[width=1.0\linewidth]
  {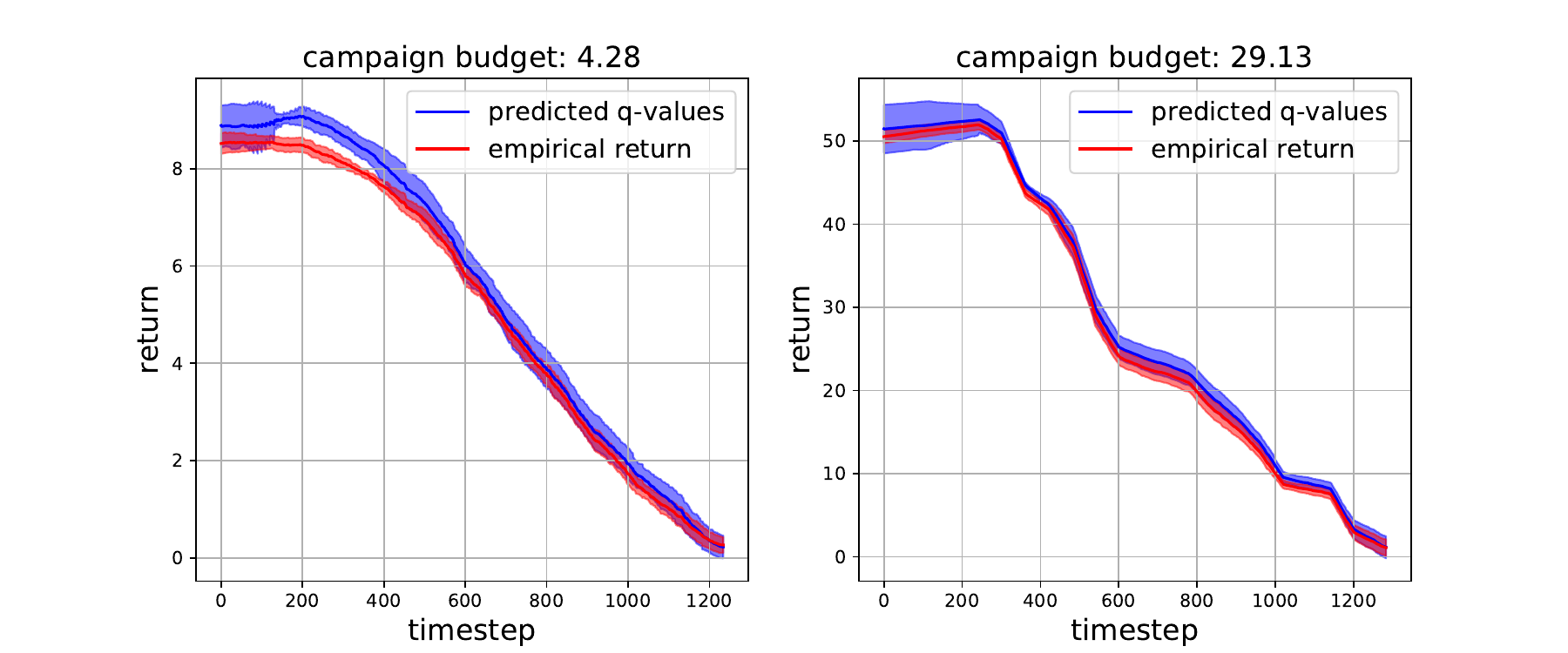}
  \caption{Learned $Q$-function predictions and empirical discounted returns-to-go, evaluated at each time step, averaged across 10 episodes for each campaign. Solid lines represent mean values and the shaded areas represent standard deviations for each curve.}
  \label{fig:q_vals_timesteps}
\end{figure}

We also wanted to verify that the critic network has learned to predict returns accurately at the intermediate states and to capture the dependence between the return, remaining budget, and remaining time in the campaign. Figure \ref{fig:q_vals_timesteps} shows predicted q-values and empirical returns evaluated at each time step for two different campaigns with different budgets. The curves are averaged across 10 episodes for each campaign, the solid lines represent the mean and the shaded areas represent the standard deviations. As can be seen, the critic model is able to accurately predict future returns at intermediate states within an episode.

Overall, these results indicate that our approach indeed can reliably improve the base policy, and, under certain algorithm hyper-parameter values, it results in a stable and consistent learning process. In particular, it is possible to confidently choose a model with a positive performance gain without evaluating multiple model checkpoints. Having verified these hypotheses in the simulator, we applied our approach to a real production system. The next section describes the details of our production setup and the results.




\subsection{Production system}\label{sec:prod_system}

\begin{table}[t]
\caption{Learning algorithm hyper-parameters values.}
\label{table:param_vals}
\begin{center}
\begin{tabular}{ |l|r| } 
\toprule
 Parameter & Value\\ 
 \midrule
 number of gradient steps during training & 600,000 \\
 \rowcolor{Gray}
 batch size & 20,000 \\ 
 critic learning rate & 3e-4 \\
  \rowcolor{Gray}
 actor learning rate & 3e-5 \\ 
  discount factor $\gamma $ & 0.9998 \\
  \rowcolor{Gray}
behavior policy st.d. $\sigma_\beta$ & 0.05 \\
CQL penalty weight $\alpha$ & 0.1 \\
\rowcolor{Gray}
exploration noise clipping range & (-0.5, 0.5) \\
critic network hidden sizes & (512, 512) \\
\rowcolor{Gray}
variance network hidden sizes & (512, 512) \\
target network update rate $\tau$ & 0.01 \\

\rowcolor{Gray}
number of samples $K$ to estimate CQL penalty terms & 50 \\

\bottomrule
\end{tabular}
\end{center}
\end{table}

We logged production data from $\sim 200,000$ week-long ad campaigns, which corresponds to $\sim 1.2$  billion time steps. One time step represents one bidding decision (computed every minute for each campaign). The data was generated by running the behavior policy defined in equation (\ref{eq:exp_noise}) directly in the production environment on a small percentage of traffic. For each recorded campaign, full episodes with all intermediate time steps were logged. The episodes terminated in one of three conditions: the campaign depleted its budget; the campaign ran for a full duration of one week; the advertiser decided to end the campaign early.  


 

\begin{table}[b] 
\caption{A/B tests results in a real production system comparing trained RL policy against the base production bidding policy. The performance metric measured in our system is a proxy to the aggregated value of impressions described in Section \ref{sec:problem_formulation}.}
\label{table:ab_tests}
\begin{center}
\begin{tabular}{ |l|l|l|}  
\toprule
 Test type  & Perf. metric gain & Impressions \\
 \midrule
pre-test  & +0.17\%, (+0.05\%, +0.3\%) 95\%CI & $\sim 50$ billion \\
back-test  & +0.16\%, (+0.03\%, +0.27\%) 95\%CI & $\sim 50$ billion \\
\bottomrule
\end{tabular}
\end{center}
\end{table}

We trained the CQL agent described in Sections \ref{sec:agent_arch} and \ref{sec:learning_algorithm} on the collected production data. After conducting a hyper-parameters search in the simulator, we used the following parameter values to train the agent (Table \ref{table:param_vals}).  
We deployed the trained parameters $w^*$ of the base policy part in our hybrid agent architecture and evaluated them online in a series of A/B tests. While we were restricted in the number of experiments and ablations we could conduct by the availability of A/B test slots that are shared between many teams, we were able to test the final model in two large-scale A/B tests.
Similarly to our results in simulation, we observed statistically significant positive performance metric gains compared to the control production bidding base policy. Each A/B test ran for one week and compared the policy learned by RL agent (test variant) to the original base policy (control variant). Table 1 shows the results of the two tests, a pre-test (a typical A/B test where a new model being tested serves as a test version) and a back-test (performed after a new model launch, where the new model serves as a control, and an old baseline now serves as a test version). In both tests, the RL-tuned policy showed statistically significant gains in the main performance metric and revenue.

\section{Conclusions}
In this work, we proposed an offline reinforcement learning approach to training parameters of production bidding policies. Auto-bidding is a natural sequential decision problem and can be formulated as a Markov decision process. Our approach allows for optimizing existing production policies from real data by directly maximizing the performance metric of interest. Our approach does not incur additional infrastructure, safety, or explainability costs, as it directly optimizes parameters of existing production routines without replacing them with black box-style models like neural networks. To validate the approach, we conducted learning experiments in both, simulated and real production bidding environments. In the simulation, we showed that our approach exhibits a consistent and stable learning performance, reliably improving upon a base bidding policy. Subsequently, we applied our approach to train the agent from the data collected in a real production environment. A series of A/B tests demonstrated statistically significant performance gains of the trained policy compared to the base policy. Based on these results we conclude that the proposed approach is a viable and practical tool for optimizing control policies under production environment constraints. We note, that, while abiding by these constraints makes our approach practically attractive, it also imposes limitations on potential performance gains we can expect. First, learning directly the parameters of a base production policy restricts us to the set of policies representable with that parameterization, and hence limits optimizing power. Second, offline RL can only reliably improve upon a base policy as long as the distribution shift is not large and the trained policy does not deviate too much from the initial policy. Overcoming these limitations in real production environments poses interesting research and engineering problems and exciting directions for future work.  


\bibliographystyle{ACM-Reference-Format}
\bibliography{rl4bidding_refs}

\end{document}